\documentclass[letterpaper, 10 pt, conference]{ieeeconf}  

\let\labelindent\relax

\usepackage[caption=false,font=scriptsize]{subfig}
\usepackage{tikz}
\usepackage{standalone}
\usepackage{times}
\usepackage{epsfig}
\usepackage{graphicx}
\usepackage{amsmath}
\usepackage{amssymb}
\usepackage{booktabs}
\usepackage[separate-uncertainty=true,multi-part-units=single]{siunitx}
\usepackage{mathtools}
\usepackage{array}
\usepackage{multirow}
\usepackage{import}
\usepackage{placeins}
\usepackage[inline]{enumitem}
\usepackage{threeparttable}
\usepackage{pgfplots}
\usepackage{layouts}
\usepackage{pdfpages}
\usepackage[space]{cite}
\makeatletter
\let\NAT@parse\undefined
\makeatother
\usepackage[hidelinks]{hyperref}

\definecolor{set2orange}{RGB}{252,141,98}
\definecolor{dark2teal}{RGB}{27,158,119}

\newlength\simheight
\pgfplotsset{compat=1.17}

\usetikzlibrary{calc}
\usetikzlibrary{positioning}
\usepgfplotslibrary{groupplots}

\IEEEoverridecommandlockouts                              

\DeclarePairedDelimiterX{\infdivx}[2]{(}{)}{%
  #1\;\delimsize\|\;#2%
}
\newcommand{\kldiv}{D_{\text{KL}}\infdivx}

\title{\LARGE \bf
Conditional Variational Autoencoders for Probabilistic Pose Regression
}

\author{Fereidoon Zangeneh$^{1,2}$, Leonard Bruns$^{1}$, Amit Dekel$^{2}$, Alessandro Pieropan$^{2}$ and Patric Jensfelt$^{1}$
\thanks{* This work was partially supported by the Wallenberg AI, Autonomous Systems and Software Program (WASP) funded by the Knut and Alice Wallenberg Foundation.}
\thanks{$^{1}$ Authors are with the division for Robotics, Perception and Learning, KTH Royal Institute of Technology, SE-100\,44 Stockholm, Sweden.
        {\tt\small \{fzk,leonardb,patric\}@kth.se}}%
\thanks{$^{2}$ Authors are with Univrses AB, SE-120\,32 Stockholm, Sweden.
        {\tt\small \{firstname.lastname\}@univrses.com}}%
}

\begin{document}

\maketitle
\thispagestyle{empty}
\pagestyle{empty}

\begin{abstract}
Robots rely on visual relocalization to estimate their pose from camera images when they lose track. One of the challenges in visual relocalization is repetitive structures in the operation environment of the robot. This calls for probabilistic methods that support multiple hypotheses for robot's pose. We propose such a probabilistic method to predict the posterior distribution of camera poses given an observed image. Our proposed training strategy results in a generative model of camera poses given an image, which can be used to draw samples from the pose posterior distribution. Our method is streamlined and well-founded in theory and outperforms existing methods on localization in presence of ambiguities.

\end{abstract}

\section{Introduction}
Localization is one of the key capabilities that robots rely on for navigation. It is the task of estimating a robot's position and orientation in its operation environment from its sensor readings. Visual localization refers to the family of methods performing this task using image observations seen by an onboard camera. It comprises techniques for both frame-to-frame visual localization, where the relative pose between two camera views is estimated, as well as global visual relocalization, where the pose for a camera view is estimated with respect to the map. An effective global relocalization technique is essential for the operation of a robot when it does not have a prior estimation of its pose, such as at start-up, when it loses track, or in the case of a kidnapped robot.

The essence of global visual relocalization, hereafter referred to as visual relocalization, is maintaining a map representation that is most apt for retrieving the pose of a novel image observation. A variety of approaches have been explored to address this task, traditionally ranging from creating a database of past image observations and comparing new observations with them \cite{torii201524, arandjelovic2016netvlad, hausler2021patch}, to building a map of salient point features in the scene and performing structure-based point matching to estimate the camera pose \cite{sattler2016efficient, liu2017efficient, sattler2012improving}. More recently a trend of works explored storing representations of the scene in weights of a neural network that directly predicts the pose of a query image \cite{kendall2015posenet, walch2017image, chen2021direct}. 

\begin{figure}[t]
    \centering
    \input{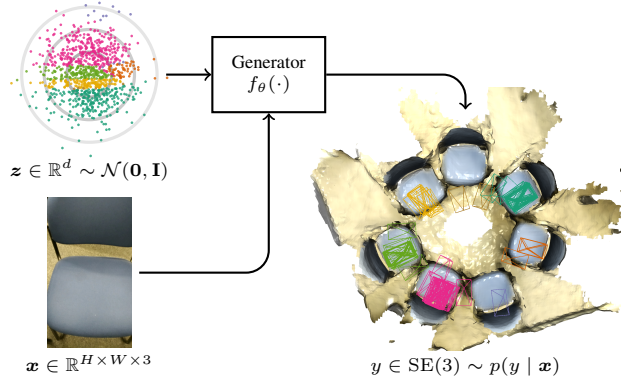}
    \caption{Our proposed generative model takes in as input a query image $\boldsymbol{x} \in \mathbb{R}^{H \times W \times 3}$ as well as a random sample $\boldsymbol{z} \in \mathbb{R}^d$ drawn from the multivariate standard Gaussian distribution, and generates a sample $y \in \mathrm{SE}(3)$ from the posterior distribution of camera poses $p(y \mid \boldsymbol{x})$. Different regions of $\mathbb{R}^d$ are mapped to distinct modes in $\mathrm{SE}(3)$, color-coded for visualization.}\label{fig:generative-model}
\end{figure}

The majority of these developed methods focus on finding the best match for a query image, and there exist several works that achieve high localization accuracy when such a single best match exists \cite{sattler2016efficient, sarlin2019coarse, moreau2022lens}. However, a scenario that raises a challenge for general visual relocalization methods is when the operation environment is inherently ambiguous, that is due to its repetitive structures, distinct camera poses record the same observation. Examples of such ambiguities include stairs in a staircase, similar chairs in a meeting room as shown in Fig. \ref{fig:generative-model}, office doors in a corridor, or machine-fabricated panels of a ceiling. The existing probabilistic approaches that can accommodate multiple hypotheses for the camera pose either rely on predicting a mixture model \cite{deng2022deep} or learn a sampling-based mapping from the space of observed visual features to camera poses within the scene \cite{zangeneh2023vapor}. However, they require some level of prior knowledge about the number of modes present in the target probability distribution, rendering them more of ad hoc solutions rather than rigorous and general approaches.

In this work we propose a novel approach with theoretical grounding for learning the conditional distribution of camera poses in a scene given a query image. Our method learns the space of camera pose ambiguities based on the visual appearances viewed in the scene, solely from a set of image and camera pose label pairs. We make a conscious choice of following the learning-based pose regression paradigm, bearing in mind its limitations in accuracy and generalization \cite{sattler2019understanding} that call for targeted remedies \cite{moreau2022lens}. We show that absolute pose regression provides a solid basis for our principled framework for probabilistic visual relocalization using conditional variational autoencoders.

In summary:
\begin{enumerate*} [label=(\arabic*)]
    \item We propose a principled approach to probabilistic visual relocalization with a generative model that predicts samples from the posterior distribution of camera poses given an image.
    \item We propose a training strategy for this generative model to learn the space of camera pose ambiguities in a scene using conditional variational autoencoders.
    \item We perform a thorough evaluation to show that our method performs better than existing solutions for handling visual ambiguities.
\end{enumerate*}

\section{Related Work}

\subsection{Visual relocalization}

Estimation of camera pose given an image is a long-standing problem in computer vision \cite{se2005vision, schindler2007city, cummins2008fab}. It is traditionally tackled by recognition-based \cite{arandjelovic2016netvlad, torii201524} and structure-based \cite{sattler2016efficient, taira2018inloc} matching solutions, with the former group known for their efficiency and scalability, and the latter for higher accuracy of their solutions. Recently, learning-based regression methods have gained traction as a promising approach for visual relocalization. These methods train neural networks that directly solve for the pose of a camera in the scene given an image. They demand smaller memory and compute resources than the traditional methods and promise higher robustness in the face of illumination changes, motion blur and texture-less surfaces \cite{walch2017image}.

One variant of regression methods is scene coordinate regression, originally proposed for RGB-D camera pose estimation \cite{shotton2013scene} and later extended to RGB data \cite{brachmann2018learning, brachmann2021visual}. These methods predict the scene's 3D point coordinates for image patches, from which the camera pose can be retrieved using a robust estimator. Although scene coordinate regression achieves similar accuracy levels as structure-based methods, end-to-end training of such pipelines for RGB data demands careful handling. A simpler approach of similar spirit is absolute pose regression, first proposed by Kendall et al.\ \cite{kendall2015posenet}, where the camera pose is directly regressed from an image. This is achieved in a pipeline with a pretrained feature extractor, whose features are mapped to the absolute camera pose in the scene by a small multi-layer perceptron. Multiple improvements to the basic idea have been subsequently investigated, for example, the use of more carefully-crafted loss functions \cite{kendall2017geometric, brahmbhatt2018geometry} and alternative architectural choices \cite{melekhov2017image, walch2017image, wang2020atloc, shavit2021learning}.

As discussed by Sattler et al.\ \cite{sattler2019understanding}, absolute pose regression in its basic form is mainly comparable to pose approximation through image retrieval. Its generalization performance to novel views falls short in comparison to image retrieval and subsequent relative geometric pose estimation via feature-matching, and consistently underperforms compared to scene coordinate regression \cite{brachmann2021visual}. Various approaches have been proposed to alleviate this downside. One promising direction is to improve the generalization by synthesizing additional, novel, views from the training data. This idea has been employed for RGB-D data \cite{naseer2017deep}, and for RGB data via estimated depth maps \cite{ng2021reassessing} and NeRFs \cite{moreau2022lens}. Other directions to improve the generalization include, for example, the use of equivariant features \cite{musallam2022leveraging} and added photometric consistency constraints on the synthesized views of predictions \cite{chen2021direct, chen2022dfnet}. In this work we base our proposed method on the absolute pose regression paradigm for its easier end-to-end training compared to scene coordinate regression. Additionally we incorporate data augmentation using NeRFs similar to \cite{moreau2022lens} in one of our experiments. However, our proposed method specifically addresses the challenge of ambiguous scenes, making it orthogonal to the aforementioned advances.

\begin{figure}[t]
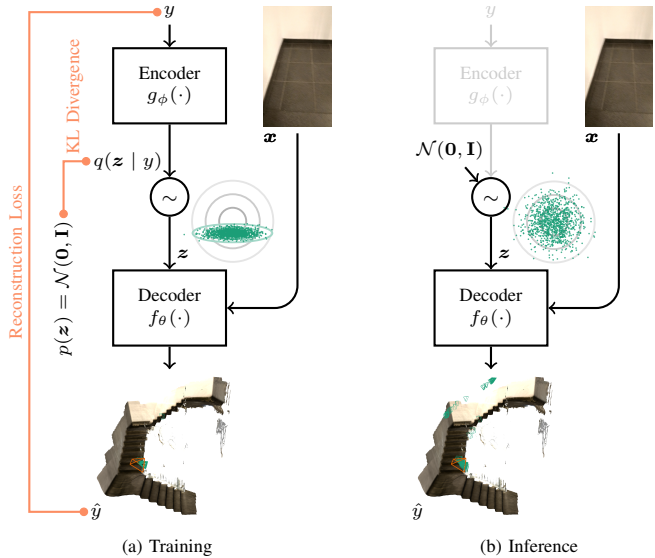

    \centering
    \subfloat[Training]{\input{figures/fig2a-portrait}}\hfill\subfloat[Inference]{\input{figures/fig2b-portrait}}
    \caption{
    (a) Our pose generative model is trained as the decoder in a conditional variational autoencoder pipeline reconstructing the ground-truth pose $y \in \mathrm{SE}(3)$ for an image $\boldsymbol{x} \in \mathbb{R}^{H \times W \times 3}$. The loss terms used in the learning objective are shown in \textcolor{set2orange}{orange}. During training the latent posterior only partly overlaps with the latent prior, resulting in generated pose samples concentrated at the ground-truth pose. (b) At inference time latent samples are drawn from the prior distribution and mapped to distinct modes in $\mathrm{SE}(3)$. In the 3D rendering of the scene we can see that for the query image viewing an ambiguous landing at the staircase, output pose samples are concentrated at three modes looking at different, but visually similar landings, including the ground truth. Pose samples are shown by \textcolor{dark2teal}{teal} and the ground truth by \textcolor{set2orange}{orange} camera frusta.}\label{fig:vae-training}
\end{figure}

\subsection{Uncertainty estimation}

A number of works propose estimating measures of uncertainty for the predictions. These works range from estimating epistemic uncertainty by training an ensemble of networks\cite{kendall2016modelling}, to estimating homoscedastic \cite{kendall2017geometric} and heteroscedastic \cite{moreau2022coordinet} aleatoric uncertainty by predicting a parametric distribution instead of a point prediction. Specifically focusing on multimodal distributions occurring for ambiguous images, Deng et al.\ \cite{deng2022deep} propose to parameterize the pose distribution as a mixture of joint Gaussian-Bingham distributions. Here, the Gaussian part describes the distribution of the position and the Bingham part describes the orientation distribution. To encourage multimodal distributions a winner-takes-all training scheme is used, in which only the mixture component closest to the target is supervised. This scheme has also been previously employed for object pose estimation \cite{fu2021multi}. With the goal of removing the explicit pose distribution parametrization, Zangeneh et al.\ \cite{zangeneh2023vapor} propose a variational inference framework in which an encoder predicts a latent distribution, which is subsequently decoded into a camera pose. Inspired by \cite{deng2022deep}, a winner\emph{s}-take-all scheme is employed. These approaches require to provide the number of expected modes in advance; in \cite{deng2022deep} the number of mixture components, and in \cite{zangeneh2023vapor} the percentage of ``winners'' has to be specified. In this work, we propose a novel conditional variational autoencoder formulation, that does not include a comparable maximum or prior on the number of modes.

Conditional variational autoencoders have been used for other vision tasks to handle multimodal target distributions. In one of the first works that promoted the use of a conditional variational autoencoder framework, Sohn et al. \cite{sohn2015learning} showcased generation of handwritten digits given a partial observation. Other works include, for example, forecasting dense pixel trajectories for static images\cite{walker2016uncertain} and image-to-image translation for domain transfer \cite{zhu2017toward}. We take inspiration from these works to formulate a principled solution to handle ambiguous scenes in visual relocalization.

\section{Method}

For visual relocalization in presence of visual ambiguities, given an image $\boldsymbol{x} \in \mathbb{R}^{H \times W \times 3}$, we are interested in estimating its posterior distribution over all camera poses $p(y \mid \boldsymbol{x})$, where $y \in \mathrm{SE}(3)$. We outline our representation and approach for estimating this distribution in Section \ref{sec:generative-model}. We then lay down the training scheme for our learned approach in Section \ref{sec:vae-training}. We finally include a short discussion on the intuition behind our method in Section \ref{sec:intuition}.

\subsection{Pose generative model} \label{sec:generative-model}

We propose to train a neural network $f_\theta(\cdot)$ that given an input image $\boldsymbol{x} \in \mathbb{R}^{H \times W \times 3}$, generates a sample $y \in \mathrm{SE}(3)$ from its pose posterior distribution. The source of randomness for this pose generator is the multivariate standard Gaussian distribution, from which a set of random samples $\mathcal{Z} =\{\boldsymbol{z}_j \sim \mathcal{N}(\mathbf{0}, \mathbf{I}) \mid j=1, \ldots, M\}$  can be drawn and fed as input to generate an arbitrarily-large set of pose samples $\mathcal{Y} =\{y_j = f_\theta(\boldsymbol{z}_j, \boldsymbol{x}) \mid \boldsymbol{z}_j \in \mathcal{Z}\}$ that represents $p(y \mid \boldsymbol{x})$. The generator network, depicted in Fig. \ref{fig:generative-model}, can be considered a learned random variable transformation of $\boldsymbol{z} \in \mathbb{R}^d$ to camera pose $y$, conditioned on the query image $\boldsymbol{x}$.

In order for the generative model to produce meaningful samples that represent $p(y \mid \boldsymbol{x})$, it must learn an appropriate mapping of $\mathbb{R}^d$ to $\mathrm{SE}(3)$, such that it transforms densities of $\mathcal{N}(\mathbf{0}, \mathbf{I})$ to $p(y \mid \boldsymbol{x})$. In other words, $\mathbb{R}^d$ is a latent space, where different regions map to the various modes in $\mathrm{SE}(3)$ in the case of a multimodal $p(y \mid \boldsymbol{x})$. This means the generator relies on an organization of the latent space, such that, given an image, latent samples that generate pose samples around the same mode in $\mathrm{SE}(3)$ are clustered together, and all samples collectively are distributed according to $\mathcal{N}(\mathbf{0}, \mathbf{I})$. We propose to learn such an organization of the latent space for all observations that the generator can be conditioned on through a conditional variational autoencoder pipeline.

\subsection{Training as a conditional variational autoencoder} \label{sec:vae-training}
We train the pose generative model as the conditional decoder of a variational autoencoder reconstructing pose samples, shown in Fig. \ref{fig:vae-training}(a). We refer the reader to \cite{doersch2016tutorial, kingma2019introduction} for a thorough introduction to variational autoencoders, and include a short summary here for completeness.

\subsubsection{Setting}
The training process of the generative model assumes a dataset of training images with known camera poses taken within the scene $\mathcal{D} = \{(\boldsymbol{x}_i, y_i) \mid i=1, \ldots, N\}$.

\subsubsection{Learning-by-reconstruction}
We frame the learning of the relation between camera poses and observations within the scene as a reconstruction task through the generative model. This reconstruction pipeline, shown in Fig. \ref{fig:vae-training}(a), consists of an inference network (encoder) $g_\phi(\cdot)$ with a Gaussian inference model, and the generative model (decoder) $f_\theta(\cdot)$. For an input pose $y_i$ the encoder predicts the mean and covariance of the Gaussian posterior distribution in the latent space $q(\boldsymbol{z} \mid y_i)$, which we can easily draw samples from. The decoder, conditioned on the observed image $\boldsymbol{x}_i$, attempts to reconstruct the original pose $y_i$ from a latent sample $\boldsymbol{z}_j \sim q(\boldsymbol{z} \mid y_i)$. This pipeline describes a variational autoencoder of camera poses conditioned on images. Following the general training scheme of variational autoencoders, the intractable true posterior distribution in the latent space $p(\boldsymbol{z} \mid y_i)$ is approximated by a class of known distributions (in this case Gaussian), and the per-pose inference of $q(\boldsymbol{z} \mid y_i)$ is amortized by training the encoder to directly predict the distribution parameters.

\subsubsection{Optimization objective terms}
The variational autoencoder pipeline is trained by maximizing the evidence lower bound (ELBO) \cite{kingma2019introduction} that consists of two terms: the Kullback-Leibler divergence $\kldiv{q(\boldsymbol{z} \mid y)}{p(\boldsymbol{z})}$ between the posterior $q(\boldsymbol{z} \mid y_i)$ and the prior distribution $p(\boldsymbol{z}) = \mathcal{N}(\mathbf{0}, \mathbf{I})$ in the latent space, and the expected log-likelihood of reconstructed samples $\mathbb{E}_{q(\boldsymbol{z} \mid y)}[\log p(y \mid \boldsymbol{z}, \boldsymbol{x})]$. The latter is estimated by Monte Carlo simulation of the posterior $\boldsymbol{z}_j \sim q(\boldsymbol{z} \mid y_i)$ to predict $\hat{y}_{i,j} = f_\theta(\boldsymbol{z}_j, \boldsymbol{x}_i)$, and compute a reconstruction loss $d_\text{pose}(\hat{y}_{i,j}, y_i)$. The optimization objective then becomes
\begin{equation}\label{eq:elbo}
    \begin{split}
        \min_{\theta, \phi} \sum_{\boldsymbol{x}_i, y_i \in \mathcal{D}} \Bigg[&\beta~\kldiv{q_\phi(\boldsymbol{z} \mid y_i)}{p(\boldsymbol{z})} \\& + \frac{1}{|\mathcal{Z}_i|}\sum_{\boldsymbol{z}_j \in \mathcal{Z}_i}d_{\text{pose}}(f_\theta(\boldsymbol{z}_j, \boldsymbol{x}_i), y_i)\Bigg],
    \end{split}
\end{equation}
where $\mathcal{Z}_i =\{\boldsymbol{z}_j \sim q_\phi(\boldsymbol{z} \mid y_i) \mid j=1, \ldots, M\}$ is the set of latent posterior Monte Carlo samples, $|\mathcal{Z}_i|$ its cardinality, and $\beta$ is the weight to balance the two terms.

\subsection{Intuition} \label{sec:intuition}

During the optimization laid out in \eqref{eq:elbo} the encoder learns to organize the camera poses in the latent space, such that the predicted latent posteriors $q(\boldsymbol{z} \mid y)$ of camera poses that view similar visual appearances are sufficiently different from each other, while collectively conforming to the standard Gaussian prior. This means that at inference time, samples from the standard Gaussian prior overlap with the latent posteriors of all camera poses that viewed a similar visual appearance during training, and as a result are mapped to various modes in $\mathrm{SE}(3)$. The mapping of different regions in the latent space to various modes in $\mathrm{SE}(3)$ is shown in Fig. \ref{fig:generative-model}, and the difference between training and inference time is further illustrated in Fig. \ref{fig:vae-training}. This clustering of camera poses given an image is learned without any prior on the number of modes or shape of the target distribution. This resolves the limitations of the existing works closest to our method, where \cite{deng2022deep} predicts a mixture model requiring the maximum number of modes to be explicitly set in advance, and \cite{zangeneh2023vapor} relies on a winners-take-all scheme to learn multimodal distributions, governed by a hyperparameter $\alpha$ that implicitly affects the number of modes that can be modeled.

\section{Implementation Details}

\subsection{Network architecture}
\emph{The generative model} (decoder) consists of a ResNet-18 backbone (with last layer set to identity) to extract $512$-dimensional image features, as well as a multilayer perceptron. Image features and the $d$-dimensional latent features are each mapped by a linear layer (+ReLU) to the common dimensionality of $64$, added, and fed as input to a fully connected network of five linear layers (+ReLU) of $128$ features. This network predicts a $3$-vector for the translation component together with a $6$-vector for the rotation component of $\mathrm{SE}(3)$. The choice of rotation representation follows the work of Zhou et al \cite{zhou2019continuity} proposing a continuous representation for regression by neural networks.

\emph{The inference network} (encoder), only used at training time, is a fully connected network with five linear layers (+ReLU) of $128$ feature, which takes in a $9$-dimensional pose sample (with the same representation as the generative model's output) and predicts the $d$-dimensional mean $\boldsymbol{\mu}$ and the log-variance $\log\boldsymbol{\sigma}^2$ that define the posterior distribution in the latent space $q(\boldsymbol{z} \mid y) = \mathcal{N}(\boldsymbol{\mu}, \text{diag}(\boldsymbol{\sigma}^2))$.

\subsection{Training setup}

We train the variational autoencoder pipeline using the Adam optimizer with a learning rate of $1 \times 10^{-4}$ and a decoupled weight decay of $1 \times 10^{-3}$ \cite{loshchilov2017decoupled}. Following prior works \cite{kendall2015posenet, deng2022deep, zangeneh2023vapor} and for improved generalization with respect to lighting changes and motion blur we apply color and brightness jittering as well as Gaussian blur on the training images, followed by a random crop of $224 \times 224$ (after they are resized such that the smallest edge is $256$ pixels). We define the reconstruction loss as
\begin{equation}
    d_{\text{pose}}([\hat{\mathbf{R}}\mid\hat{\boldsymbol{t}}], [\mathbf{R}\mid\boldsymbol{t}]) = \lambda_r \lVert \hat{\mathbf{R}} - \mathbf{R} \rVert_\mathrm{F} + \lambda_t \lVert \hat{\boldsymbol{t}} - \boldsymbol{t} \rVert_2,
\end{equation}
with $\lambda_r$ and $\lambda_t$ as balancing weights tuned to reflect the metric size of the scenes. We weigh the KL divergence term by $\beta=0.3$ with a warm-up period of $4000$ starting after $1000$ iterations. We define the latent space to be $4$-dimensional and train with a batch size of $16$ in all experiments. At both training and test time we draw $1000$ latent samples to represent the camera pose posterior for each image.

\section{Experiments}

\subsection{Datasets and scenarios}
We evaluate our method in three settings, two from state-of-the-art works for benchmarking \cite{zangeneh2023vapor, deng2022deep}, extended by one synthetic scenario for a controlled analysis of our method.

\subsubsection{Real-life ambiguous scenes}
We evaluate our method on the Ambiguous Relocalization sequences \cite{deng2022deep} as well as a sequence of a moving camera facing a ceiling from \cite{zangeneh2023vapor}, which captures a scenario of more severe ambiguity. We use these sequences to benchmark our method against the prior works. As the sequences in Ambiguous Relocalization dataset are short, for better generalization and following \cite{moreau2022lens} we augment each training set by training a NeRF on the training samples \cite{mueller2022instant} and synthesizing novel views uniformly-spaced along the camera trajectory, perturbed by Gaussian noise with $\sigma = 10\si{\cm}$ and $10^\circ$. We perform this for our method as well as the baseline \cite{zangeneh2023vapor}.

\subsubsection{Real-life unambiguous scenes}
We also evaluate our method on the visual relocalization datasets 7-Scenes \cite{shotton2013scene} and Cambridge Landmarks \cite{kendall2015posenet} to show its performance on general visual relocalization task in natural scenes.

\subsubsection{Synthetic ambiguities}
In order to examine the central thesis of our proposed method and decouple it from disturbances to the learning task---lighting changes, motion blur and dataset imbalance, we create a synthetic scenario, where a camera observes a scene made up of three distinct colors. The images are solid colors and are viewed in succession at a steady rate of camera movement. This creates a case of complete ambiguity to a learned network, without any seemingly benign but ultimately unambiguous speckles or details that tend to exist in real-life data. We use this synthetic sequence to showcase the functionality of our proposed method in a completely controlled and simplified environment.

\subsection{Evaluation protocol}

\begin{figure*}[t]
    \centering
    \input{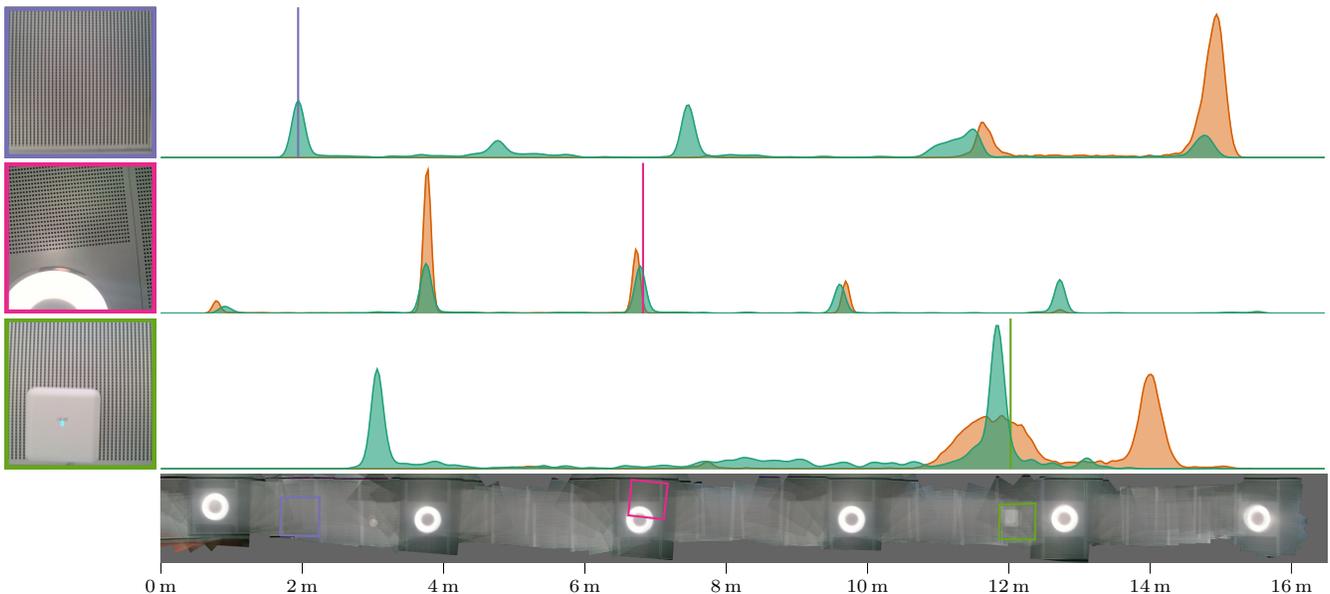}
    \caption{Predicted marginal posterior distributions for the ceiling scene. Each row depicts an example, with the query image on the left, and predictions of our method in \textcolor{dark2teal}{teal} and VaPoR \cite{zangeneh2023vapor} in \textcolor{set2orange}{orange} on the right. We represent each distribution with $1000$ samples and show the marginalized distribution along the longitudinal translation axis. For visualization we show the results of kernel density estimation on the predicted samples. The ground-truth position in each example is shown by a vertical solid line, color-coded with the border of the query image and its projection on the ceiling. We can see that our method produces better predictions; in the first example VaPoR fails to assign density at the ground-truth position, and in the third example we deem the prediction of our method more coherent, predicting density at the position of the viewed WiFi router, as well as the smoke detector on the ceiling.}\label{fig:ceiling}
\end{figure*}

The true pose posterior for an ambiguous query image is by definition non-unimodal and has probability density also at regions distinct from the ground-truth pose label. For this reason, following \cite{zangeneh2023vapor} we measure the quality of a predicted pose posterior distribution by its recall. For a query image, we count its predicted pose posterior distribution as true positive if it contains sufficient density in the vicinity of the ground-truth pose label, and a false negative otherwise. In our sample-based representation of the posterior we assess the sufficiency of density by checking if at least a ratio $\gamma$ of all samples are within a threshold of distance to the ground-truth pose. The appropriate value of $\gamma$ depends on the degree of inherent ambiguity in the scene. Following \cite{zangeneh2023vapor} we use $\gamma = 0.1$ for Ambiguous Relocalization sequences and $\gamma = 0.05$ for the more ambiguous ceiling sequence.

Existing works on the task of visual relocalization commonly assess accuracy through median translation and orientation errors \cite{kendall2015posenet, brahmbhatt2018geometry, kendall2016modelling}. We adopt this metric to evaluate our method on unambiguous scenes. To do so we obtain point estimates for the predicted distributions by computing arithmetic and chordal $L_2$ \cite{hartley2013rotation} means of the translation and rotation components of their samples, respectively, followed by median error calculation across the test set.

\begin{table}[t]
    \centering
    \begin{threeparttable}
    \caption{Measured recall in ambiguous scenes for thresholds $0.1\si{\meter} / 10^{\circ}$, $0.2\si{\meter} / 15^{\circ}$, $0.3\si{\meter} / 20^{\circ}$ (higher is better)}\label{tab:ambiguous}
    \scriptsize
    \setlength{\tabcolsep}{3.0pt}
    \begin{tabular}{@{}lcccc@{}}
        \toprule
        \# & MN \cite{brahmbhatt2018geometry} & BMDN \cite{deng2022deep} & VaPoR \cite{zangeneh2023vapor} & Ours\\
        \midrule
        1 & $0.05 / 0.33 / 0.56$ & $0.41 / 0.83 / 0.89$ & $0.72 / \mathbf{1.00} / \mathbf{1.00}$ & $\mathbf{0.79} / 0.96 / 0.96$\\
        2 & $0.00 / 0.03 / 0.07$ & $0.09 / 0.27 / 0.33$ & $0.05 / 0.43 / \mathbf{0.61}$ & $\mathbf{0.15} / \mathbf{0.45} / 0.53$\\
        3 & $0.07 / 0.17 / 0.29$ & $0.24 / 0.48 / 0.69$ & $0.19 / \mathbf{0.61} / \mathbf{0.78}$ & $\mathbf{0.29} / 0.58 / 0.75$\\
        4 & $0.01 / 0.03 / 0.07$ & $0.11 / 0.43 / 0.60$ & $0.01 / 0.19 / 0.42$ & $\mathbf{0.15} / \mathbf{0.48} / \mathbf{0.62}$\\
        5 & $0.09 / 0.37 / 0.53$ & $0.38 / 0.79 / 0.91$ & $0.45 / 0.86 / 0.93$ & $\mathbf{0.52} / \mathbf{0.92} / \mathbf{0.98}$\\
        \midrule
        6 & $0.02 / 0.05 / 0.09$ & $0.08 / 0.19 / 0.30$ & $0.33 / 0.60 / 0.68$ & $\mathbf{0.53} / \mathbf{0.73} / \mathbf{0.81}$\\
        \bottomrule
    \end{tabular}
    \begin{tablenotes}
         Scenes:
         \begin{enumerate*} [label=(\arabic*)]
            \item Blue Chairs (Fig. \ref{fig:generative-model}),
            \item Meeting Table,
            \item Staircase (Fig. \ref{fig:vae-training}),
            \item Staircase Extended,
            \item Seminar,
            \item Ceiling (Fig. \ref{fig:ceiling}).
        \end{enumerate*}
        \vspace{-0.2cm}
    \end{tablenotes}
    \end{threeparttable}
\end{table}

\section{Results and Discussion}

\subsection{Real-life ambiguous scenes}

We compare our method to Deep Bingham Networks \cite{deng2022deep} (marked BMDN) and VaPoR \cite{zangeneh2023vapor}, looking at the recall values of each across the ambiguous scenes. While capable of handling ambiguities in the scene, these two reference probabilistic pose regression methods, at least implicitly, rely on prior knowledge regarding the maximum number of modes in the target distribution. We evaluate BMDN with 10 mixture components accommodating a maximum of 10 modes, and VaPoR with the implicit parameter $\alpha=0.05$ and a decoder depth of $5$, giving both methods sufficient capacity to handle the ambiguities of the test scenes. We identified and rectified an implementation error in VaPoR's codebase, resulting in improved performance compared to their original recall values reported in \cite{zangeneh2023vapor}. In our evaluation we also include MapNet \cite{brahmbhatt2018geometry} (marked MN) as an effective end-to-end absolute pose regression method to highlight the need for probabilistic approaches for handling ambiguities.

\begin{table}[t]
    \centering
    \begin{threeparttable}
        \caption{Median error ($\si{\meter} / ^{\circ}$) in unambiguous scenes (lower is better)}\label{tab:unambiguous}
        \scriptsize
        \setlength{\tabcolsep}{4.0pt}
        \begin{tabular}{@{}lccccc@{}}
            \toprule
            Scene &  MN$\smash{^*}$ \cite{brahmbhatt2018geometry} & BPN$\smash{^*}$ \cite{kendall2016modelling} &  BMDN$\smash{^*}$ \cite{deng2022deep} & VaPoR \cite{zangeneh2023vapor} & Ours\\
            \midrule
            Chess & $\mathbf{0.08/3.25}$ & $0.37/7.24$ & $0.10/6.47$ & $0.15/11.8$ & $0.15/7.04$\\
            Fire & $0.27/\mathbf{11.7}$ & $0.43/13.7$ & $\mathbf{0.26}/14.8$ & $0.32/15.7$ & $0.30/13.6$\\
            Heads & $0.18/13.3$ & $0.31/\mathbf{12.0}$ & $\mathbf{0.13}/13.4$ & $0.21/16.1$ & $0.15/14.3$\\
            Office & $\mathbf{0.17/5.15}$ & $0.48/8.04$ & $0.19/9.73$ & $0.24/12.1$ & $0.24/8.99$\\
            Pumpkin & $0.22/\mathbf{4.02}$ & $0.61/7.08$ & $\mathbf{0.20}/9.40$ & $0.27/13.2$ & $0.30/8.54$\\
            Kitchen & $0.23/\mathbf{4.93}$ & $0.58/7.54$ & $\mathbf{0.19}/10.9$ & $0.25/14.2$ & $0.28/9.48$\\
            Stairs & $\mathbf{0.30/12.1}$ & $0.48/13.1$ & $0.34/14.1$ & $0.30/16.2$ & $0.36/14.5$\\
            \midrule
            College & $1.07/\mathbf{1.89}$ & $1.74/4.06$ & $1.51/2.14$ & $1.07/4.27$ & $\mathbf{1.01}/3.74$\\
            Hospital & $\mathbf{1.94/3.91}$ & $2.57/5.14$ & $2.25/3.93$ & $2.74/5.18$ & $2.59/4.35$\\
            Façade & $1.49/\mathbf{4.22}$ & $1.25/7.54$ & $3.52/5.41$ & $1.00/4.84$ & $\mathbf{0.98}/5.22$\\
            Church & $2.00/\mathbf{4.53}$ & $2.11/8.38$ & $2.16/5.99$ & $1.86/7.13$ & $\mathbf{1.53}/7.94$\\
            \bottomrule
        \end{tabular}
        \begin{tablenotes}
             \item[$*$] Results of MapNet, Bayesian PoseNet and BMDN are taken from \cite{zangeneh2023vapor}.
        \end{tablenotes}
    \end{threeparttable}
\end{table}

\begin{figure*}[t]
    \centering
    \input{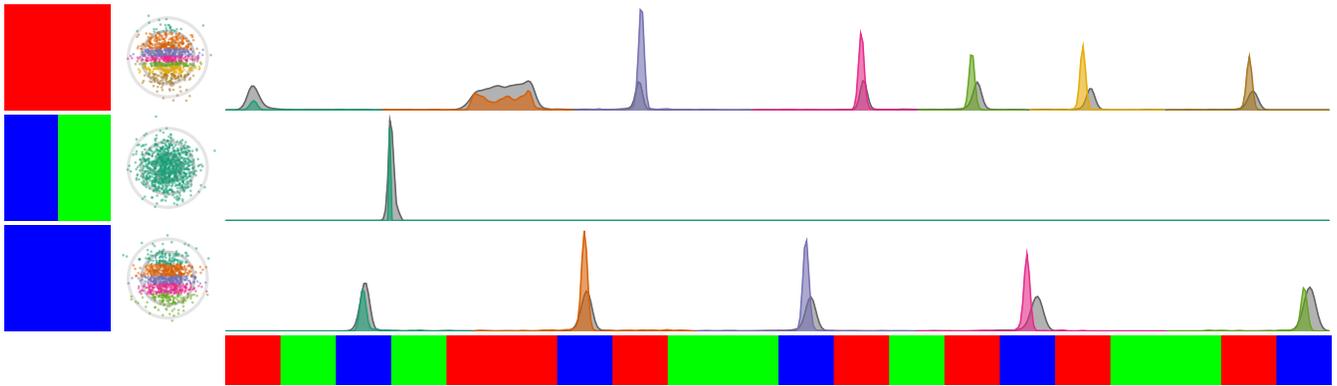}
    \caption{Predicted marginal posterior distributions for our synthetic scene. Each row depicts an example, with the query image on the far left, followed by samples from the latent space, and predicted posterior distributions on the right. We obtain each distribution through a similar procedure as in Fig. \ref{fig:ceiling}, and color-code predictions of our method together with the latent samples according to the ground-truth modes of the posterior. Predictions of VaPoR \cite{zangeneh2023vapor} are shown in \textcolor{gray}{gray}. We can see that our method can effectively learn the camera pose ambiguities for each query image, splitting the latent space into different regions mapping to distinct modes when the image is ambigous (first and third examples), while mapping the whole latent space to a single mode when the image is unambiguous (second example).}\label{fig:colors}
\end{figure*}

We can see in Table \ref{tab:ambiguous} that our proposed method has an edge on other methods, while not assuming any prior knowledge about the ambiguities in the scene. Fig. \ref{fig:ceiling} depicts three examples from the ceiling sequence, showcasing our method's more coherent predictions in comparison with VaPoR. Across different scenes we observed that our method often predicts distributions with sharper peaks compared to VaPoR, which tends to predict modes with wider spreads that are possibly favored by the adopted recall-based evaluation protocol. As also reported in \cite{zangeneh2023vapor}, we observed that although BMDN's winner-takes-all strategy effectively positions mixture components, it struggles to predict the correct mixture weights required for generating coherent distributions.

\subsection{Real-life unambiguous scenes}
For completeness and to ensure the applicability of our method in a general visual relocalization setting, we report the accuracy of our method in common unambiguous benchmark datasets in Table \ref{tab:unambiguous}. The table also includes the Bayesian PoseNet \cite{kendall2016modelling}, another well-known absolute pose regression method. We see that our approach, even with the additional machinery of a sampling-based probabilistic method, achieves comparable performance to methods streamlined to predict a single accurate solution. This shows that our proposed framework does not impose an additional limitation on the capabilities of absolute pose regression.

\subsection{Synthetic ambiguities}

Fig. \ref{fig:colors} illustrates our synthetic ambiguous sequence and highlights the relation between the latent space and the predicted posterior distribution for three example cases. In this controlled experiment, we know the true locations of the modes in the posterior distribution. So, we employ color-coding to associate the predicted posterior with the true modes, together with the samples drawn from the latent prior. We can see that the decoder can effectively learn the latent space of pose ambiguities given an image. That is, it splits the latent space into different regions depending on the ambiguity of the query image, and maps each to a distinct mode in $\mathrm{SE}(3)$.

\subsection{Run-time analysis}
We measure the time it takes to generate $1000$ samples from the pose posterior distribution for a query image, corresponding to a forward pass of the generative model. With 100 repetitions, this on average takes $8.50\pm0.65\si{\ms}$ on CPU (Intel Core i9-13900KF), which means our method can run in real time.

\section{Future Work}
 We experimented with normalizing flows \cite{rezende2015variational} for more flexible latent posterior shapes than Gaussian. However, we decided to exclude it as we did not observe consistent performance improvement across different scenes as a result of it. We hypothesize that in our setting the space of camera pose ambiguities could be modeled sufficiently well by low-dimensional Gaussian posteriors. However, future work could revisit the use of normalizing flows for likelihood estimation of pose samples \cite{kolotouros2021probabilistic}. Another promising direction is the use of diffusion models, which have shown remarkable generative capabilities in other tasks \cite{gong2023diffpose, zhang2024generative, shrestha2023caldiff}. They can supplant the variational autoencoder setup for handling multimodal target distributions in visual relocalization.

\section{Conclusion}
In this work we revisit the problem of visual relocalization in the face of ambiguities and propose a novel approach with theoretical grounding and derived from first principles. At the core of our method is the learning of the space of camera pose ambiguities for image observations in a scene. We show that this can be materialized in a conditional variational autoencoder pipeline, yielding a generative model that given an image produces samples from its camera pose posterior distribution. Unlike the existing works, our proposed approach does not assume any prior knowledge about the shape of the target distribution. We perform an extensive evaluation to examine the working of our method and compare it to existing methods, showing that our approach has a performance edge on other works in ambiguous scenes.


{\small
\bibliographystyle{IEEEtran}
\bibliography{bib}
}

\end{document}